\documentclass[10pt,twocolumn,letterpaper]{article}

\usepackage{iccv}
\usepackage{times}
\usepackage{epsfig}
\usepackage{graphicx}
\usepackage{amsmath}
\usepackage{amssymb}
\usepackage{subfigure}
\usepackage{float}
\usepackage{multirow}

\graphicspath{{pics/}}

\usepackage[pagebackref=true,breaklinks=true,letterpaper=true,colorlinks,bookmarks=false]{hyperref}

\iccvfinalcopy 

\def\iccvPaperID{7430} 

\ificcvfinal\fi

\begin{document}

\title{RPVNet: A Deep and Efficient Range-Point-Voxel Fusion Network for LiDAR Point Cloud Segmentation}


\author{
Jianyun Xu$^1$\thanks{Equal contribution. The first two authors are listed in the alphabetical order.~$^\dag$Corresponding author.}
\quad
Ruixiang Zhang$^{1,2}$\footnotemark[1]
\quad
Jian Dou$^1$
\quad
Yushi Zhu$^1$
\\
Jie Sun$^1$
\quad
Shiliang Pu$^{1\dag}$
\\
$^1$Hikvision Research Institute \quad $^2$ Zhejiang University
\\
{\tt\small \{xujianyun, zhangruixiang7, doujian, zhuyushi, sunjie, pushiliang.hri\}@hikvision.com}
}



\maketitle
\ificcvfinal\thispagestyle{empty}\fi

\begin{abstract}
   Point clouds can be represented in many forms (views), typically, point-based sets, voxel-based cells or range-based images(i.e., panoramic view). The point-based view is geometrically accurate, but it is disordered, which makes it difficult to find local neighbors efficiently. The voxel-based view is regular, but sparse, and computation grows cubicly when voxel resolution increases. The range-based view is regular and generally dense, however spherical projection makes physical dimensions distorted. Both voxel- and range-based views suffer from quantization loss, especially for voxels when facing large-scale scenes. In order to utilize different view's advantages and alleviate their own shortcomings in fine-grained segmentation task, we propose a novel \textbf{r}ange-\textbf{p}oint-\textbf{v}oxel fusion network, namely RPVNet. In this network, we devise a deep fusion framework with multiple and mutual information interactions among these three views, and propose a \textbf{g}ated \textbf{f}usion \textbf{m}odule (termed as GFM), which can adaptively merge the three features based on concurrent inputs. Moreover, the proposed RPV interaction mechanism is highly efficient, and we summarize it to a more general formulation. By leveraging this efficient interaction and relatively lower voxel resolution, our method is also proved to be more efficient. Finally, we evaluated the proposed model on two large-scale datasets, i.e., SemanticKITTI and nuScenes, and it shows state-of-the-art performance on both of them. Note that, our method currently ranks \textbf{1st} on SemanticKITTI leaderboard without any extra tricks.
\end{abstract}


\begin{figure}[t]
\centering
\subfigure[Point-based: disordered]{
\includegraphics[width=0.9\linewidth]{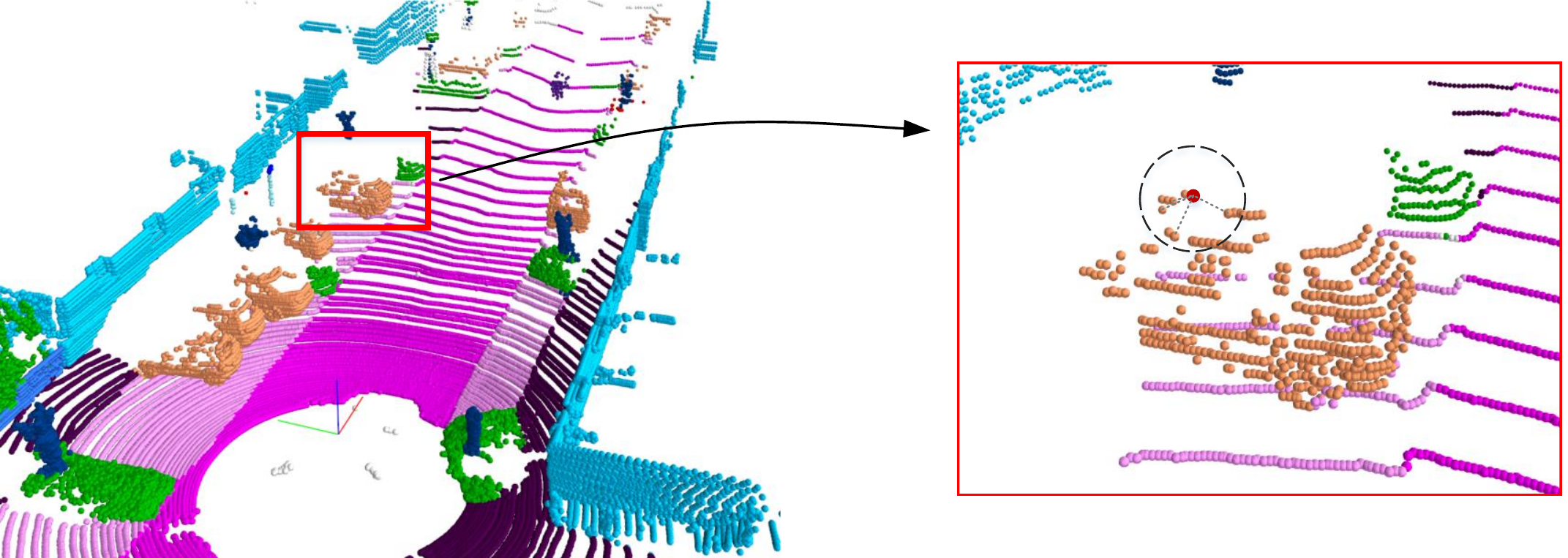}
\label{fig.subfig.rpv-p}
}
\quad
\subfigure[Voxel-based: sparse, quantization loss]{
\includegraphics[width=0.9\linewidth]{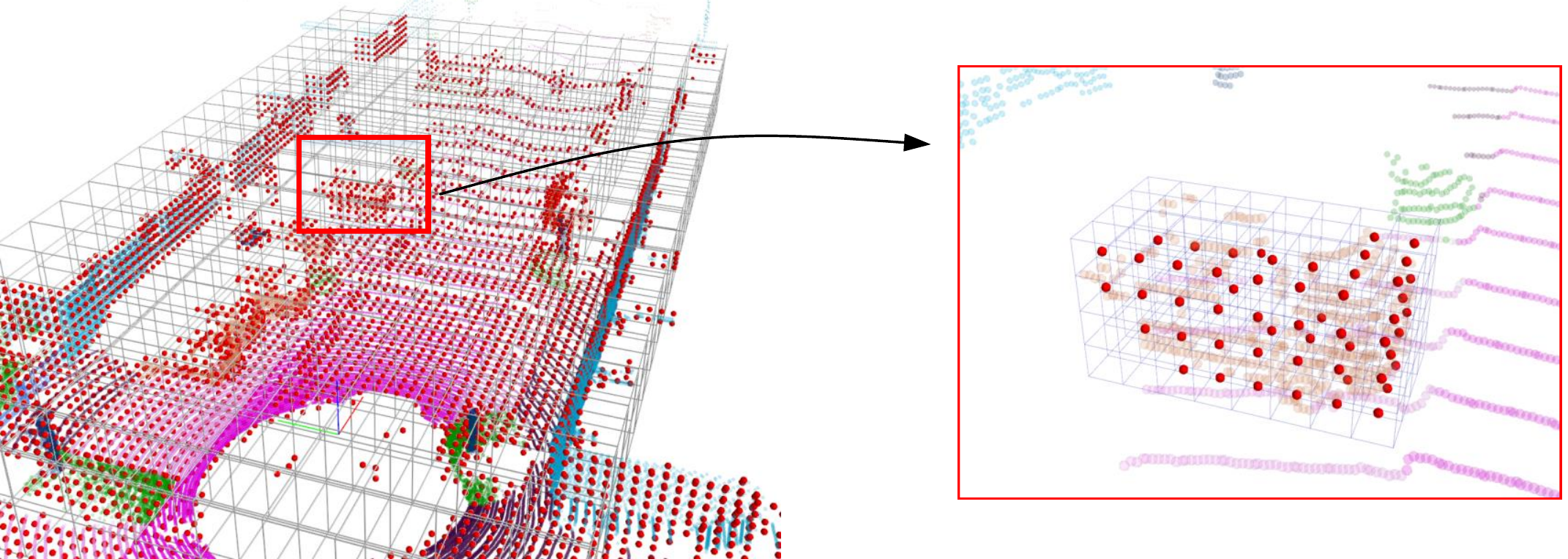}
\label{fig.subfig.rpv-v}
}
\quad
\subfigure[Range-based: physical dimensions distorted]{
\includegraphics[width=0.9\linewidth]{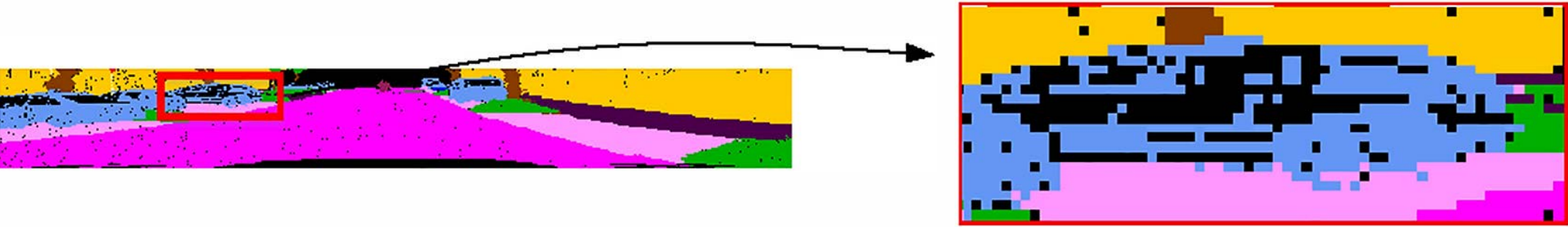}
\label{fig.subfig.rpv-r}
}
\caption{All three views have shortcomings. (a) point-based: the points are irregular, which makes finding the neighbors of a point inefficient. (b) voxel-based: voxelization brings quantization loss, and the computation grows cubicly when resolution increases. (c) range-based: range-image distorts physical dimensions because of spherical projection. }
\label{fig.rpv_show}
\end{figure}

\section{Introduction}
3D computer vision is receiving more and more attention due to its wide range of applications, such as AR/VR, robotics and autonomous driving. In this paper, we aim to improve the performance of semantic segmentation in driving scenario, so as to provide high-quality point-wise perception of the entire 3D scene.  Collected by the Light Detection and Ranging (LiDAR) sensors, 3D data usually comes in the format of point clouds. And there are several common forms (views) to represent it along with some specific preprocesses.

Conventionally, researchers rasterize the point cloud into voxel cells, as depicted in Figure~\ref{fig.subfig.rpv-v}, and process them using 3D volumetric convolutions~\cite{choy20163d, riegler2017octnet}. With voxels defined based on the Cartesian coordinate system, voxel-based view retains physical dimensions and has a friendly memory locality. Nevertheless, it is relatively sparse and requires very high resolution in order to get rid of quantized information loss, which brings a cubic increase in both computation and memory footprint.

Recently, the point-based view, as shown in Figure~\ref{fig.subfig.rpv-p}, has attracted more and more attention, and various works tend to consume points directly. PointNet~\cite{qi2016pointnet} is the pioneer work which uses per-point Multi-Layer Perceptions(MLPs) to extract point features, but it lacks the local context modeling capability. Based on PointNet, later researches~\cite{qi2017pointnetplusplus, pcnn, wu2019pointconv, wang2019dgcnn, thomas2019kpconv} have paid extensive attention to each point's local feature extraction by aggregating its neighboring features. However, points are unstructured, thus inefficient to search one point's neighbors due to random memory access.

A parallel track of works~\cite{wu2018squeezeseg, xu2020squeezesegv3, iros19rangenetplusplus, cortinhal2020salsanext} follows a spherical projection scheme, i.e., range-based view, as shown in Figure~\ref{fig.subfig.rpv-r}, where the sub-spaces of 3D information in the form of depth map are learned with well studied 2D networks. In these approaches, a convolution tower on range-image can aggregate information across a large receptive field, helping to alleviate the point sparsity issue. Due to the spherical projection, however, physical dimensions are not distance-invariant and objects may overlap each other severely in a cluttered scene.

\begin{figure}[t]
\centering

\subfigure[methods vs mIoU]{
    \begin{minipage}{0.47\linewidth}

    \includegraphics[width=1.1\linewidth]{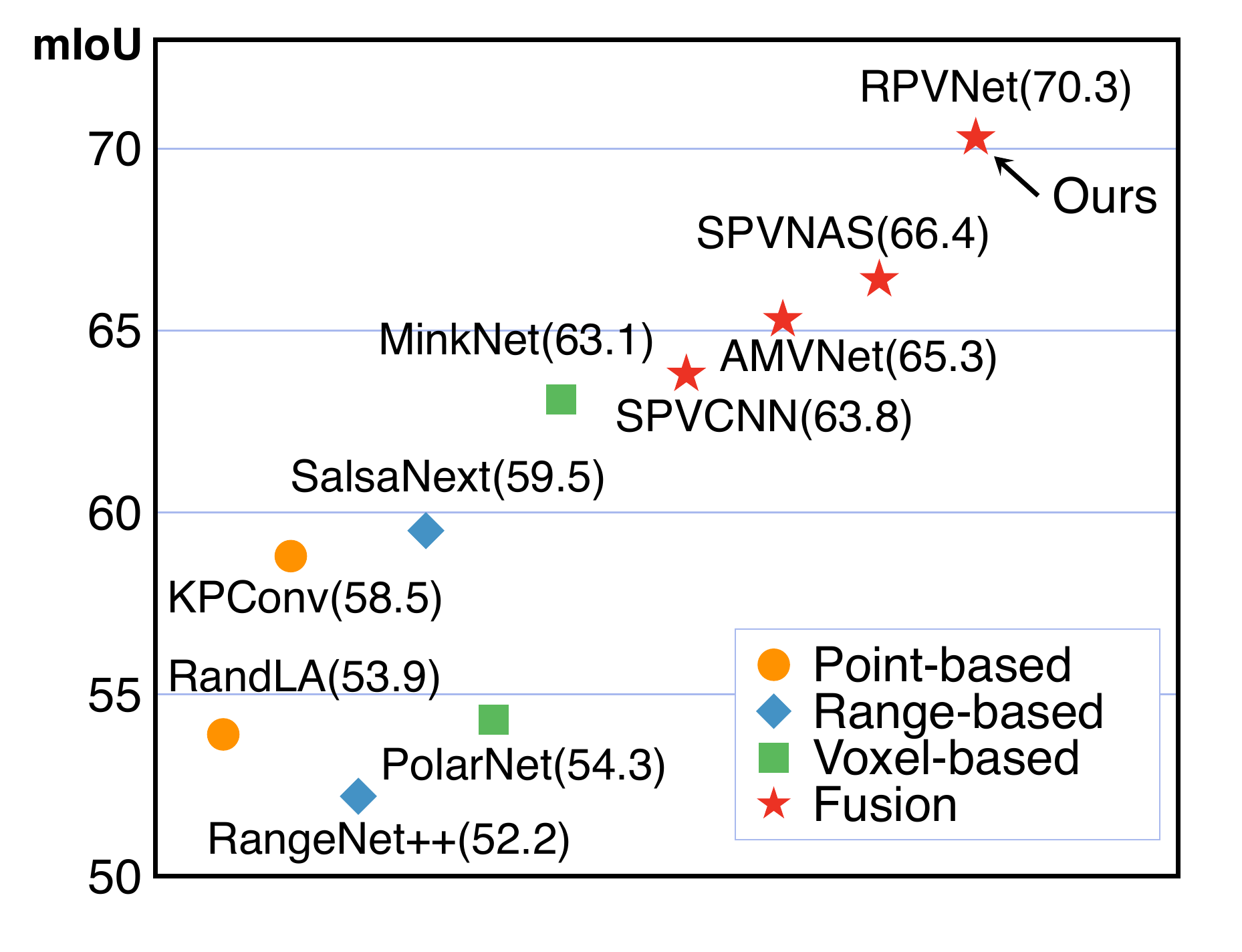}
    \label{fig.subfig.performance}
    \end{minipage}
}
\subfigure[voxel resolution vs mIoU]{

    \begin{minipage}{0.47\linewidth}

    \includegraphics[width=1.1\linewidth]{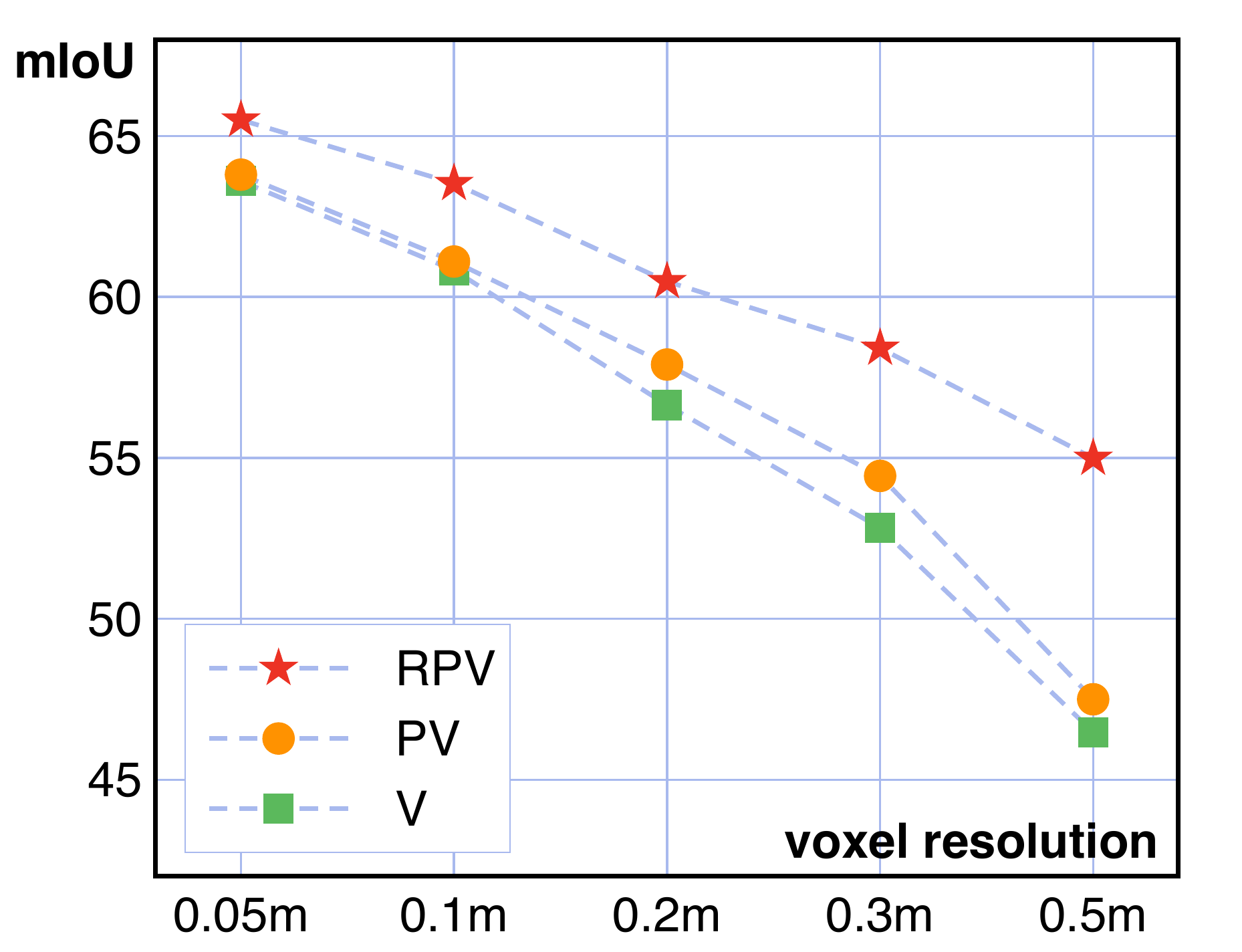}
    \label{fig.subfig.vres-miou}
    \end{minipage}

}




\caption{Performance overview. (a) performance distribution of different methods. It indicates that voxel-based methods is generally better than point- and range-based methods, and multi-view fusion based method is better than single-view methods. (b) performance versus voxel resolution. V denotes voxel-based method, PV is point-voxel fusion method, and RPV is our range-point-voxel fusion method. We can discover that PV fusion can only get a limited improvement, but our RPV fusion can consistently boost the performance even when the voxel resolution is relatively high.}
\label{fig.miou}
\end{figure}

In aspect of point cloud segmentation in large-scale driving scene, we discover that: 1). voxel-based methods are relatively higher in performance than point- and range-based methods, meanwhile the best ones of point- and range-based methods are basically the same, as depicted in Figure~\ref{fig.subfig.performance}; 2). range-based methods are relatively more efficient than point- and voxel-based methods because of highly optimized 2D convolution, and point-based methods are far from real-time requirements when involving local neighbors searching; 3). voxel-based methods are hard to keep a high voxel resolution when efficiency is also taken into account, and the performance drop sharply when resolution decreases, as shown in Figure~\ref{fig.subfig.vres-miou}. 

It is intuitive to combine different views together, leveraging the complementary information while preserving advantages and alleviating shortcomings. One recent attempt is PVCNN~\cite{liu2019pvcnn}, a point-voxel fusion scheme, in which voxel offers a coarse-grained local feature, while point preserves fine-grained geometrical features by running simple per-point MLPs. It provides a great perspective, but the performance improvement brought by point-voxel fusion is limited, as depicted in Figure~\ref{fig.subfig.vres-miou}, and it is not sufficient to use a simple additive fusion. 

In this paper, we propose a deep and adaptive range-point-voxel fusion framework aiming to synergize all the three views' representations. More specifically, as shown in Figure~\ref{fig:fusion}, we design a fusion strategy using points as middle hosts, and transfer features on range-pixels and voxel-cells to points, then apply an adaptive feature selection in order to choose the best feature representation for each point, and finally transfer the fused features on points back to range-image and voxels. Compared to other previous multi-view fusion methods~\cite{cvpr17-mv3d, e2e-mvf, pillar-based-mvf, liong2020amvnet, zhang2020deepfusionnet, gerdzhev2020tornado}, which either fuse in the front or the end of network, our method conducts aforementioned fusion multi-times in the network, which allows different views to enhance each other in a deeper and more flexible way. As for efficiency: Firstly, we propose an efficient RPV interaction mechanism by utilizing hash mapping. Secondly, we use a relatively lower voxel resolution and sparse convolution in voxel branch. Thirdly, we perform simple MLPs on point branch similar to~\cite{liu2019pvcnn}, getting rid of inefficient local neighbors searching. Finally, we employ a highly efficient range branch to decrease computation. Moreover, we find that the class is extremely imbalanced in datasets, thus we design an \textit{instance CutMix} augmentation in the training phase to alleviate the class imbalance problem.

Main contributions in this paper are listed as follows:
\begin{itemize}
    \item  We devise a deep and adaptive range-point-voxel fusion framework, which allows different views to enhance each other in a more flexible way. 
    \item  We propose an efficient RPV interaction mechanism by utilizing hash mapping, and summarize it to a more general formulation for future extension.
    \item We conduct massive experiments to evalute the effectiveness and efficiency of our proposed method, and our method achieves state-of-the-art results on both SemanticKITTI~\cite{semantickitti} and nuScenes~\cite{caesar2020nuscenes} datasets.
\end{itemize}




\section{Related Work}
\subsection{Point-based Segmentation}
PointNet~\cite{qi2016pointnet} firstly pioneered the direct operation on points through a MLP-based network. Although the subsequent elaborate works~\cite{qi2017pointnetplusplus, pcnn, wu2019pointconv, wang2019dgcnn, thomas2019kpconv} have been shown to be effective on indoor point cloud data, most of them cannot be directly scaled to large-scale outdoor data due to computational and memory limitations. RandLA-Net~\cite{hu2020randla} uses random sampling and a local feature aggregation to reduce the information loss which brought by random operation. It provides a feasible way to accelerate the point network, but it cannot escape the accuracy loss caused by sampling. KPConv~\cite{thomas2019kpconv} achieves the best results of current point-based methods with its novel spatial kernel-based point convolution. However, facing the same problem, it cannot directly train the entire data for large scenes. Compromisingly, category-balanced sampling by the radius is used to reduce the data scale. However, this division may destroy some intrinsic information of point cloud. Overall, 
despite that point-based methods may have lower parameters~\cite{hu2020randla}, but they inevitably involing the inefficient local neighbors searching.

\subsection{Voxel-based Segmentation}
The early voxel-based methods~\cite{chang2015shapenet,maturana2015voxnet,qi2016volumetric,choy20163d,riegler2017octnet,zhou2018voxelnet,wang2019voxsegnet} convert points into voxels and apply vanilla 3D convolutions for semantic segmentation. More recently, some efforts~\cite{MinkowskiNet, spvnas} have been made to accelerate the 3D convolution, and improve the performance to a higher level with less computation. Simultaneously, variants~\cite{zhang2020polarnet, zhu2020cylinder3d} of 3D space partition have also been proposed. Among them, Cylinder3D~\cite{zhu2020cylinder3d} designs an asymmetrical residual block to reduce computation and ensures features related to cuboid objects. 
AF2S3Net~\cite{cheng20212} stands on the shoulder of ~\cite{MinkowskiNet} and achieves the state-of-the-art of past methods, which proposes two novel attention blocks named Attentive Feature Fusion Module(AF2M) and Adaptive Feature Selection Module(AFSM), to effectively learn local and global contexts and emphasize the fine detailed information, moreover, a mixed loss function with geo-aware anisotropy~\cite{li2019depth} is also leveraged to recover the fine details.
When the resolution is reduced, voxel methods will suffer serious information loss, but our proposed method use other views to make up this disadvantage.

\subsection{Range-based Segmentation}
The range-based point cloud segmentation approachs~\cite{iros19rangenetplusplus,cortinhal2020salsanext,kochanov2020kprnet,wu2018squeezeseg,xu2020squeezesegv3} utilize 2D CNN by projecting 3D point clouds into 2D dense spherical grids. 
For instance,  RangeNet++~\cite{iros19rangenetplusplus} draws on the DarkNet backbone in YOLOv3~\cite{redmon2018yolov3} as the extractor and is  post-processed by using an accelerated KNN. SalsaNext~\cite{cortinhal2020salsanext} takes Salsanet~\cite{aksoy2020salsanet} as baseline and presents an uncertainty-aware mechanism for point feature learning. Otherwise, KPRNet~\cite{kochanov2020kprnet} stands out among such methods and achieves state-of-the-art results by using a strong ResNeXt-101 backbone with an Astrous Spatial Pyramid Pooling layer, meanwhile, it also uses KPConv~\cite{thomas2019kpconv} as a segmentation head to replace the inefficient KNN innovately. Although range-based methods can use mature 2D image segmentation technology, spherical projection process will distort physical dimensions, which can be 
avoid in our proposed RPVNet with multi-view interactive learning.

\subsection{Multi-View Fusion}
Since the single view is more or less problematic, some recent approaches~\cite{e2e-mvf,pillar-based-mvf,liong2020amvnet,zhang2020deepfusionnet,gerdzhev2020tornado,spvnas} attempt to blend two or more different views together. For instance, methods in~\cite{e2e-mvf,pillar-based-mvf} perform an early-fusion by combining point-level information from bird-eye-view and range-image before feeding it to the subsequent network. AMVNet~\cite{liong2020amvnet} designs a late-fusion method by calculating the uncertainty of different views' output and uses an extra network to refine the results. FusionNet~\cite{zhang2020deepfusionnet} proposes a point-voxel interaction MLP that aggregates features between neighborhood voxels and corresponding points, which reduces time consuming in neighbor search and achieves acceptable accuracy on the large-scale point cloud. 
In particular, PVCNN~\cite{liu2019pvcnn} propose an efficient point-voxel fusion scheme, in which voxel offers a coarse-grained local feature, while point preserves fine-grained geometrical features by running simple per-point MLPs. The above methods only use two views, besides fusion method between them is very simple(eg.addition). However, our proposed method can make more effective use of multi perspective information, and select and fuse useful parts.

\begin{figure*}
    \centering
    \includegraphics[width=0.93\linewidth]{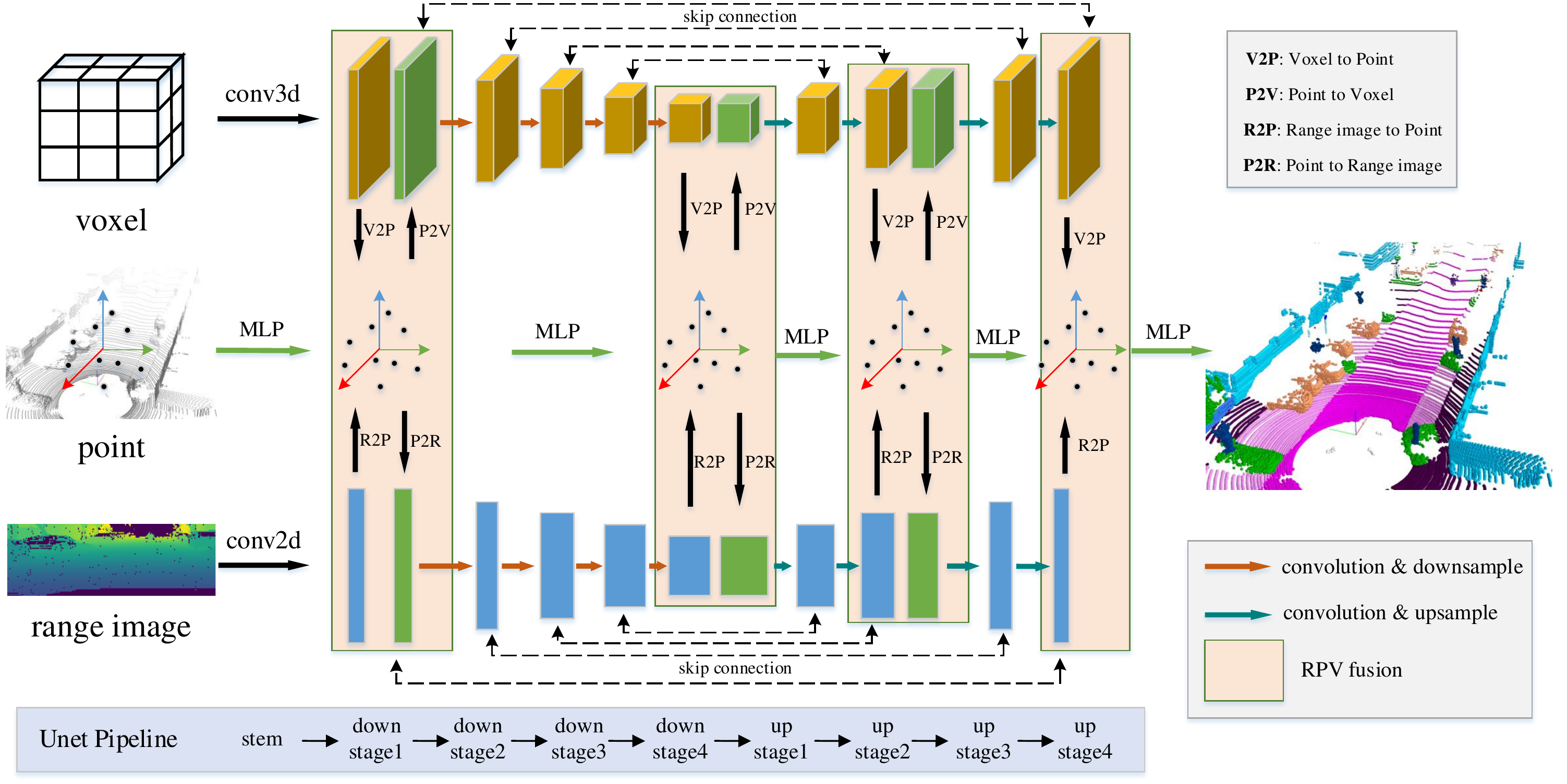}
    \caption{Overview of RPVNet. It is a three-branch network with multiple interactions among them, where voxel- and range-branch share the similar Unet architecture, and point-branch only utilize per-point MLPs.}
    \label{fig:framework}
\end{figure*}

\section{Methodology}
\label{sec:Methodology}
In this Section we first present the overview of our network, then introduce its crucial components, i.e., RPV interaction and fusion methods. Finally, an effective data augmentation we adopted is also introduced.

\subsection{Framework Overview}
\label{sec:Framework}
The block diagram of the proposed method, RPVNet, is illustrated in Figure~\ref{fig:framework}. It is a three-branch network with multiple interactions among them. From top to down, the three branches are voxel-, point- and range-branch, respectively. We use a very similar Unet for both voxel- and range-branch. The Unet structure first uses a stem to extract contextual information from the original input, then performs four down-sampling stages, and finally connects four up-sampling stages to restore the original points. Point-branch is an extremely simple PointNet~\cite{qi2016pointnet} with several MLPs. RPV-fusion happens after the stem, the fourth down-sampling, the second up-sampling and the last up-sampling stage, which is the same as SPVCNN~\cite{spvnas} for fair comparison. The details of multi-view interactions will be illustrated in the following section.

\subsection{Efficient Multi-View Interactive Learning} 

The original point cloud can be converted into different views, typically voxels and range-images, so points can be used as an intermediate carrier to build connections among these views, any form of point cloud representation can be seen as a mapping of the original points. Therefore, we achieve efficient multi-view interactive learning by building multi-view representation indexing and multi-view feature propagation. The unified feature mapping and representation of multi view are realized by 
indexing system. Multi perspective feature interaction and learning are acquired through feature propagation. Consequently, we build range-point-voxel interacting module in our proposed RPVNet, which will be detailed below.

{\bf Multi-View Representation Indexing.}
Given a point cloud in the form of points as $P\in \mathbb{R}^{N\times (3+C)}$, we can transform it to any other form $X\in \mathbb{R}^{M\times D}$ through some ``projection'' function, where $N$ denotes the number of points, $3$ indicates its $xyz$ coordinates in Euclidean space, C is the point's feature channels , $M$ denotes the number of elements in the form $X$, and $D$ is the dimension indicating its positional information. To construct the connection between $P$ and $X$, we first use the ``projection'' function $\mathcal{P}: \mathbb{R}^{N\times (3+C)}\mapsto \mathbb{R}^{M\times D}$, subsequently followed by a hash function $\mathcal{H}:\mathbb{R}^{M\times D}\mapsto \mathbb{N}^{M}$. Thus we can build a hash mapping from $P$ to $X$, which is a highly efficient search. We denote $X$ as an arbitrary representation of a point cloud, and  any form of point cloud representation $X$ can be seen as a mapping of the original points $P$. To note that,``projection'' function $\mathcal{P}$ may be a many-to-one mapping, hash function $\mathcal{H}$ is a one-to-one mapping in our case. 

{\bf Multi-View Feature Propagation.} 
The feature of element $j$ in other form of point cloud representation $X$ is affected by the 
corresponding original points. Moreover, since the ``projection'' function $\mathcal{P}$ may be non-injection, $j$ will often be jointly affected by multiple keys from $P$. Define all keys of $j$ as $\mathcal{K}_X(j) = \{k,\mathcal{H}(\mathcal{P}(k))=\mathcal{H}(j)\}$,the feature of $j$ is propagated from corresponding keys. In general, the feature propagation from point to other view can be formulated as:
\begin{equation} \label{con:P2X}
    \mathcal{F}_X(j)=\Omega(j,\mathcal{F}_P)
\end{equation}
where $\mathcal{F}_X \in \mathbb{R}^{M\times D}$,$\mathcal{F}_P \in \mathbb{R}^{N\times (3+C)} $ is the feature of $X$ and $P$, $\Omega(\cdot)$ denotes a weighting function(eg.averaging,max).

On the contrary, we define the inverse feature propagation function as $\Phi(\cdot)$,which reflects the flow of information from other view to point view. The formula is as follow:
\begin{equation} \label{con:X2P_a}
    \mathcal{F}_P(i)=\Phi(j,\mathcal{F}_X).
\end{equation}
where $i$ denotes a point in $P$.

{\bf Range-Point-Voxel Interacting}
Point-based representation preserves the geometric details and capture the fine-grained information which is friendly for some small instance(eg.\textit{pedestrian},\textit{bicycle}). Voxel-based representation maps points into a regular 3D grid while maintaining the spatial structure, which effectively extracts information in various 3D dimensions. However, the existence of voxel resolution will introduce the quantization loss. Range-based representation converts sparse points into a structured and dense image, and large receptive field brings richer semantic information for large-scale categories. To this end, we have effectively used the information of these three views, and realized interactive learning between them to improve the performance of various categories.

As mentioned above, we first establish the Range-Point-Voxel indexing system. Specifically including projection indexing and voxelization indexing from the points to the range-view  and voxel representation. Projection indexing is formulated by projecting points onto a spherical surface to generate the range-image as in ~\cite{rangeview-rangID}. With rangelization, we get an range-image $R\in\mathbb{R}^{H\times{W}\times{D}}$ with height $H$, width $W$ and the normalized coordinates $G={(n,m)}$ of the points in the map, where the normalized coordinates indicate points on a map of any scale. Apart from the mapping between point and range image, we represent the voxelization indexing through transforming the input point cloud $P$ to a sparse voxel representation $V$. For different voxel size $r$, the ``projection'' mapping is $\mathcal{P}_V:\mathcal{F}_{P(z)} \stackrel{r}{\longrightarrow{}} \mathcal{F}_{V(\lfloor z/r \rfloor )}$, where $z$ is the point coordinate. 

Based on the Range-Point-Voxel indexing system, we further perform the feature propagation of each other. In this work, we use averaging to pass the features from Point to other two views. So, we can rewrite $\Omega(j,\mathcal{F}_P)$ in Eq.\ref{con:P2X} as follows:

\begin{equation}
\Omega(j, \mathcal{F}_P)=\frac{\sum_{u \in \mathcal{K}_X(j)}\mathcal{F}_P (u)}{Num(\mathcal{K}_X(j))}
\end{equation}
where $Num(\cdot)$ denotes the count function, $u$ indicates the element of $\mathcal{K}_X(j)$.
Furthermore, the partial derivatives of features can be computed as:
\begin{equation}\label{con:X2P}
    \frac{\partial \mathcal{F}_X(j)}{\partial \mathcal{F}_p(u)} = \frac{1}{Num(\mathcal{K}_X(j))}.
\end{equation}

Moreover, for other view to point view's feature propagation function, the nearest neighbor interpolation  can be implemented as $\Phi(j,\mathcal{F}_X)=\mathcal{F}_X(\lfloor j \rfloor)$ for simplicity . Besides, we followed the \cite{spvnas} implementation which utilizes trilinear interpolation with eight-neighborhood voxel grids for voxel to point view. Similarly, our range to point view process uses the bilinear interpolation method as:
\begin{flalign}
& \mathcal{F}_P(i)=\Phi(\delta(j), \mathcal{F}_R)=\sum_{u\in\delta(j)} \phi(u,j)\mathcal{F}_R(u) \label{con:P2R}\\
& \phi(u,j)=(1-\lfloor  j_x  - u_x \rfloor)(1-\lfloor j_y - u_y  \rfloor)
\end{flalign}
where $\phi(\cdot)$  computes the bilinear weights, $\delta(j)$ denotes the four-neighbor grids of $j$. Therefore,the partial derivative of Eq.\ref{con:X2P} becomes:
\begin{equation} \label{con:R2P_g}
    \frac{\partial \mathcal{F}_P(i)}{\partial \mathcal{F}_R(u)} = \sum_{u\in\delta(j)} \phi(u,j).
\end{equation}

\begin{figure}[t]
    \centering
    \includegraphics[width=1.0\linewidth]{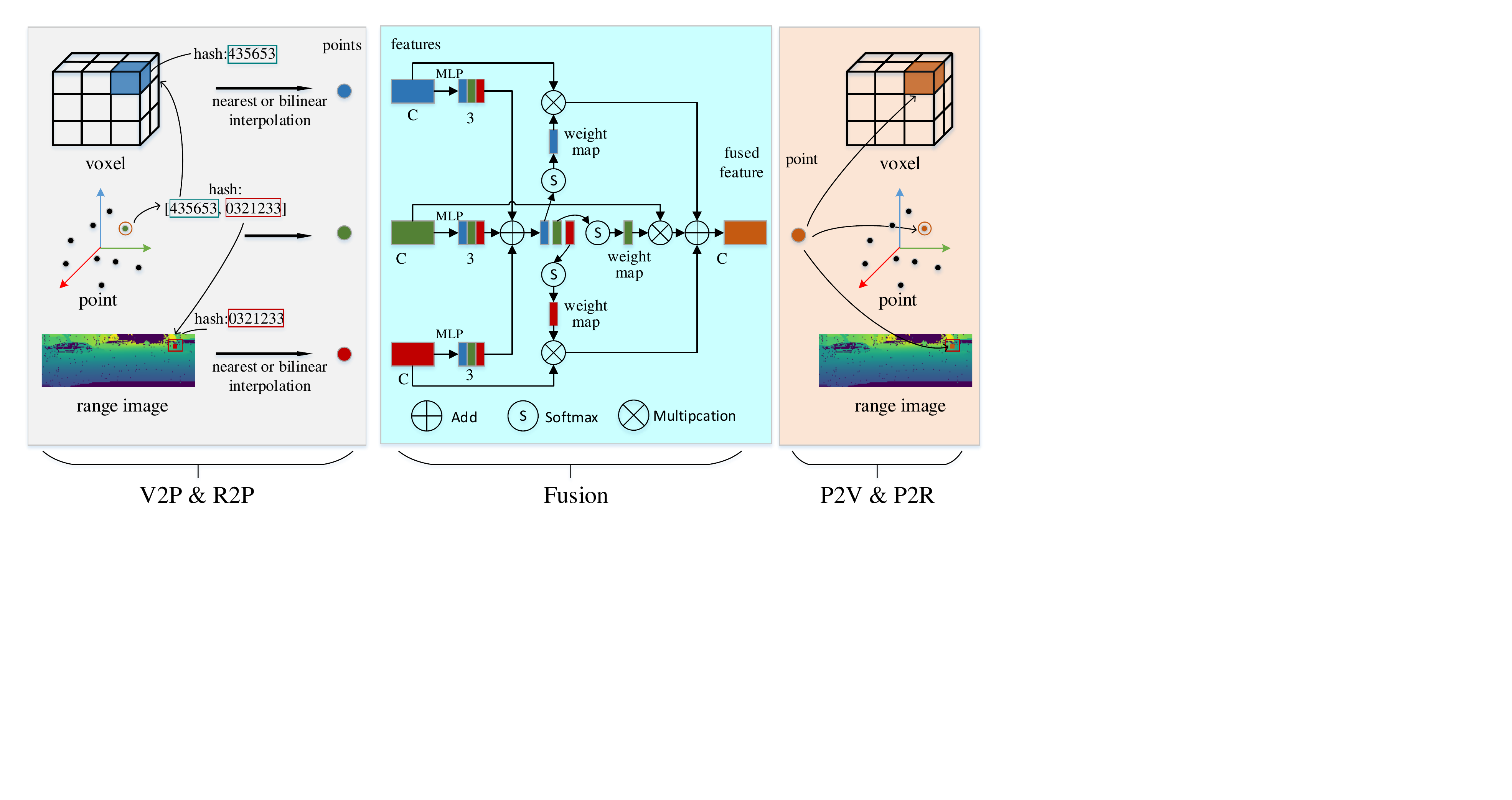}
    \caption{Details of RPV fusion. In left block: Given a point and its hash code, we need to find the corresponding voxel or pixel. In middle block: Given features on point from different views, we need to merge them adaptively. In right block: Given the fused point feature, we need to project it back to other views.}
    \label{fig:fusion}
\end{figure}


\subsection{Gated Fusion Module(GFM)}
Given $L$ feature vectors ${X_i \in \mathbb{R}^{N_i \times C_i}}$ from different view branches, $N_i$ ,$C_i$ are the number of points and channels of the $i$th feature vector respectively.
The essential task in multi-view feature fusion is to aggregate useful information together under the interference of massive useless information.  Addition and Concatenation are commonly used operations to aggregate multiple features, but both of them will be influenced by numerous non-informative features. They can be formulated as:
\begin{flalign}\label{con:cat}
\centering
& \widetilde{X}=concat(X_1,...,X_L),\\
& \widetilde{X}=\sum_{i=1}^{L} X_i,
\end{flalign}
where $\widetilde{X}$ is the fused feature vector.The features of different view branches are not consistent in importance due to their variability. But the above primary fusion strategies ignore the usefulness of each feature vector and combine massive useless features with useful features during fusion. 

Inspired by the mature Gating mechanism\cite{cho2014learning,takikawa2019gated,li2020gated} which can aggregate information adaptively by measuring the importance of each feature, our gated fusion module is designed based on the normal addition-based fusion by filtering information flow with gates.The gated fusion is formally defined as:
\begin{equation}\label{con:gfm}
    \widetilde{X}=\sum_{i}^{L} split[softmax(\sum_{i}^{L} G_i)]_i \cdot X_i
\end{equation}
where $\cdot$ denotes element-wise multiplication and $G_i\in[0,1]^{N \times L}$ means gate vector for $i$th representation.Each gate vector has $L$ channels for each representation.The feature weight votes on each channel are superimposed by summation and converted into probability weights by softmax. Finally, the consequences on the corresponding channels are separated to weight the input features. Each gate vector $G_i = sigmoid(w_i * X_i)$ is estimated by a convolutional layer parameterized with $w_i \in \mathbb{R}^{1\times{L}\times{C_i}}$.The detailed operation can be seen in Fig.\ref{fig:fusion}(b).

\begin{table*}[]
\begin{center}
\resizebox{\textwidth}{!}{%
\begin{tabular}{l|c|ccccccccccccccccccc}
\hline
  \textbf{Methods} &
  \textbf{mIoU} &
  \rotatebox{90} {car} &
  \rotatebox{90} {bicycle} &
  \rotatebox{90} {motorcycle} &
  \rotatebox{90} {truck} &
  \rotatebox{90} {other-vehicle} &
  \rotatebox{90} {person} &
  \rotatebox{90} {bicyclist} &
  \rotatebox{90} {motorcyclist} &
  \rotatebox{90} {road} &
  \rotatebox{90} {parking} &
  \rotatebox{90} {sidewalk} &
  \rotatebox{90} {other-ground} &
  \rotatebox{90} {building} &
  \rotatebox{90} {fence} &
  \rotatebox{90} {vegetation} &
  \rotatebox{90} {trunk} &
  \rotatebox{90} {terrain} &
  \rotatebox{90} {pole} &
  \rotatebox{90} {traffic-sign} \\ \hline\hline
PointNet~\cite{qi2016pointnet}        & 14.6 & 46.3 & 1.3  & 0.3  & 0.1  & 0.8  & 0.2  & 0.2  & 0.0  & 61.6 & 15.8 & 35.7 & 1.4  & 41.4 & 12.9 & 31.0 & 4.6  & 17.6 & 2.4  & 3.7  \\
RandLANet~\cite{hu2020randla}       & 53.9 & 94.2 & 26.0 & 25.8 & 40.1 & 38.9 & 49.2 & 48.2 & 7.2 & 90.7 & 60.3 & 73.7 & 20.4 & 86.9 & 56.3 & 81.4 & 61.3  & 66.8 & 49.2 & 47.7 \\
KPConv~\cite{thomas2019kpconv}          & 58.8 & 96.0 & 30.2 & 42.5 & 33.4 & 44.3 & 61.5 & 61.6 & 11.8 & 88.8 & 61.3 & 72.7 & 31.6 & 90.5 & 64.2 & 84.8 & 69.2 & 69.1 & 56.4 & 47.4 \\ \hline
SqueezeSegv3~\cite{xu2020squeezesegv3}    & 55.9 & 92.5 & 38.7 & 36.5 & 29.6 & 33.0 & 45.6 & 46.2 & 20.1 & 91.7 & 63.4 & 74.8 & 26.4 & 89.0 & 59.4 & 82.0 & 58.7 & 65.4 & 49.6 & 58.9 \\
RangeNet++~\cite{iros19rangenetplusplus}      & 52.2 & 91.4 & 25.7 & 34.4 & 25.7 & 23.0 & 38.3 & 38.8 & 4.8  & 91.8 & 65.0 & 75.2 & 27.8 & 87.4 & 58.6 & 80.5 & 55.1 & 64.6 & 47.9 & 55.9 \\
SalsaNext~\cite{cortinhal2020salsanext}       & 59.5 & 91.9 & 48.3 & 38.6 & 38.9 & 31.9 & 60.2 & 59.0 & 19.4 & 91.7 & 63.7 & 75.8 & 29.1 & 90.2 & 64.2 & 81.8 & 63.6 & 66.5 & 54.3 & 62.1 \\ \hline
PolarNet~\cite{zhang2020polarnet}        & 54.3 & 93.8 & 40.3 & 30.1 & 22.9 & 28.5 & 43.2 & 40.2 & 5.6  & 90.8 & 61.7 & 74.4 & 21.7 & 90.0 & 61.3 & 84.0 & 65.5 & 67.8 & 51.8 & 57.5 \\
MinkowskiNet~\cite{MinkowskiNet}       & 63.1$^*$ & - & - & - & - & - & - & - & - & - & - & - & - & - & - & - & - & - & - & -  \\
Cylinder3D~\cite{zhu2020cylinder3d} & 67.8 & 97.1 & 67.6 & 64.0 & \textbf{59.0} & 58.6 & 73.9 & 67.9 & 36.0 & 91.4 & 65.1 & 75.5 & 32.3 & 91.0 & 66.5 & 85.4 & 71.8 & 68.5 & 62.6 & 65.6 \\ 
AF2S3~\cite{cheng20212}           & 69.7 & 94.5 & 65.4 & \textbf{86.8} & 39.2 & 41.1 & \textbf{80.7} & \textbf{80.4} & \textbf{74.3} & 91.3 & 68.8 & 72.5 & \textbf{53.5} & 87.9 & 63.2 & 70.2 & 68.5 & 53.7 & 61.5 & \textbf{71.0} \\\hline
FusionNet~\cite{zhang2020deepfusionnet}       & 61.3 & 95.3 & 47.5 & 37.7 & 41.8 & 34.5 & 59.5 & 56.8 & 11.9 & 91.8 & 68.8 & 77.1 & 30.8 & 92.5 & 69.4 & 84.5 & 69.8 & 68.5 & 60.4 & 66.5 \\
TornadoNet~\cite{gerdzhev2020tornado}      & 63.1 & 94.2 & 55.7 & 48.1 & 40.0 & 38.2 & 63.6 & 60.1 & 34.9 & 89.7 & 66.3 & 74.5 & 28.7 & 91.3 & 65.6 & 85.6 & 67.0 & 71.5 & 58.0 & 65.9 \\
AMVNet~\cite{liong2020amvnet}          & 65.3 & 96.2 & 59.9 & 54.2 & 48.8 & 45.7 & 71.0 & 65.7 & 11.0 & 90.1 & \textbf{71.0} & 75.8 & 32.4 & 92.4 & 69.1 & 85.6 & 71.7 & 69.6 & 62.7 & 67.2 \\
SPVCNN~\cite{spvnas}          & 63.8 & - & - & - & - & - & - & - & - & - & - & - & - & - & - & - & - & - & - & - \\
SPVNAS~\cite{spvnas}          & 67.0 & 97.2 & 50.6 & 50.4 & 56.6 & 58.0 & 67.4 & 67.1 & 50.3 & 90.2 & 67.6 & 75.4 & 21.8 & 91.6 & 66.9 & 86.1 & 73.4 & 71.0 & 64.3 & 67.3 \\ \hline
\textbf{RPVNet}          & \bf{70.3}      & \bf{97.6}     & \textbf{68.4}     &68.7      &44.2      &\textbf{61.1}      &75.9      &74.4      &73.4      &\textbf{93.4}      &70.3      &\textbf{80.7}      &33.3      &\textbf{93.5}      &\textbf{72.1}      &\textbf{86.5}      &\textbf{75.1}      &\textbf{71.7}      &\textbf{64.8 }     &61.4     \\ \hline

\end{tabular}%
}
\end{center}
\caption{Class-wise and mean IOU of our proposed method and state-of-the-art methods on SemanticKITTI leaderboard. The methods are grouped as point-based, range-based, voxel-based and fusion networks. $*$: result reproduced by~\cite{spvnas}. Note that, our result uses the \textit{instance CutMix} augmentation (see Sec.~\ref{sec:cutmix}), and voxel resolution is set to 0.05m, but \textbf{without} extra tricks. Accessed on 18 March 2021.}
\label{tab:res-kiti}
\end{table*}


\begin{table*}[]
\begin{center}
\resizebox{0.9\textwidth}{!}{%
\begin{tabular}{l|c|cccccccccccccccc}
\hline
  \textbf{Methods} &
  \textbf{mIoU} &
  \rotatebox{90} {barrier} &
  \rotatebox{90} {bicycle} &
  \rotatebox{90} {bus} &
  \rotatebox{90} {car} &
  \rotatebox{90} {construction} &
  \rotatebox{90} {motorcycle} &
  \rotatebox{90} {pedestrian} &
  \rotatebox{90} {traffic-cone} &
  \rotatebox{90} {trailer} &
  \rotatebox{90} {truck} &
  \rotatebox{90} {driveable} &
  \rotatebox{90} {other\_flat} &
  \rotatebox{90} {sidewalk} &
  \rotatebox{90} {terrain} &
  \rotatebox{90} {manmade} &
  \rotatebox{90} {vegetation} \\ \hline\hline
RangeNet++~\cite{iros19rangenetplusplus} & 65.5 & 66.0 & 21.3 & 77.2 & 80.9 & 30.2 & 66.8 & 69.6 & 52.1 & 54.2 & 72.3 & 94.1 & 66.6 & 63.5 & 70.1 & 83.1 & 79.8 \\
PolarNet~\cite{zhang2020polarnet}   & 71.0 & 74.7 & 28.2 & 85.3 & 90.9 & 35.1 & 77.5 & 71.3 & 58.8 & 57.4 & 76.1 & 96.5 & 71.1 & 74.7 & 74.0 & 87.3 & 85.7 \\
Salsanext~\cite{cortinhal2020salsanext}  & 72.2 & 74.8 & 34.1 & 85.9 & 88.4 & 42.2 & 72.4 & 72.2 & 63.1 & 61.3 & 76.5 & 96.0 & 70.8 & 71.2 & 71.5 & 86.7 & 84.4 \\
AMVNet~\cite{liong2020amvnet}     & 76.1 & \textbf{79.8} & 32.4 & 82.2 & 86.4 & \textbf{62.5} & 81.9 & 75.3 & \textbf{72.3} & \textbf{83.5} & 65.1 & \textbf{97.4} & 67.0 & \textbf{78.8} & 74.6 & \textbf{90.8} & 87.9 \\
Cylindr3D~\cite{zhu2020cylinder3d}  & 76.1 & 76.4 & 40.3 & 91.2 & \textbf{93.8} & 51.3 & 78.0 & 78.9 & 64.9 & 62.1 & \textbf{84.4} & 96.8 & 71.6 & 76.4 & 75.4 & 90.5 & 87.4 \\ \hline
\textbf{RPVNet}     & \textbf{77.6} & 78.2 & \textbf{43.4} & \textbf{92.7} & 93.2 & 49.0 & \textbf{85.7} & \textbf{80.5} & 66.0 & 66.9 & 84.0 & 96.9 & \textbf{73.5} & 75.9 & \textbf{76.0} & 90.6 & \textbf{88.9} \\ \hline
\end{tabular}%
}
\end{center}
\caption{Class-wise and mean IOU of our proposed method and state-of-the-art methods on nuScenes validation set. Note that out RPVNet shows result without any tricks including \textit{instance CutMix}.}
\label{tab:res-nuscenes}
\end{table*}

\subsection{Instance CutMix}
\label{sec:cutmix}
Although some data augmentation approaches\cite{li2020pointaugment,chen2020pointmixup,zhang2021pointcutmix} for indoor point cloud have shown to be effective, there is little research on outdoor scenes. Inspired by past mix-based approaches, we proposed the instance mix to cope with the imbalanced class problem for lidar semantic segmentation.

Empirically, the network can predict less frequent objects more accurately if such objects are allowed to repeated in the scene. Motivated by this discover, we extract the rare-class object instances(eg.\textit{bicycles}, \textit{vehicles}) from each frame of the training set into a mini sample library. During the training, the samples are randomly selected from the mini-sample pool equally by category. Then, random scaling and rotation will be acted on these samples. To ensure a close fit with reality, we randomly placed the objects above the ground-class points. Finally, some new rare objects from other scenes are ``pasted'' to the current training scenes, for simulating objects in various environments.




\section{Experiments}

In this section, we introduce the implementation details of our RPVNet (Sec.~\ref{sec:Experimental Setup}) and compare with previous state-of-the-art methods on both the highly competitive SemanticKITTI dataset~\cite{semantickitti} (Sec.~\ref{sec: results_kitti}) and the newly introduced large-scale nuScenes Dataset~\cite{caesar2020nuscenes}(Sec.~\ref{sec: results_nuscenes}). Also in Sec.~\ref{sec: results_kitti}, more details about computation and learnable parameters have been presented to illustrate our method's efficiency. Finally, in Sec.~\ref{sec: Ablation Studies}, we conduct extensive ablation studies to investigate crucial components of the RPVNet, and leads to some promising conclusions.

\subsection{Experimental Setup}
\label{sec:Experimental Setup}
{\bf Datasets.}
\textit{SemanticKITTI}~\cite{semantickitti}, derived from the KITTI Vision Benchmark~\cite{kitti}, is a large-scale dataset for point cloud segmentation task in driving scene. It consists of 43551 LiDAR scans from 22 sequences collected in a city of Germany. Equipped with a Velodyne-HDLE64 LiDAR, each scan has approximately 120k points. These 22 sequences are split into 3 sets, i.e., training set (00 to 10 except 08 with 19130 scans), validation set (08 with 4071 scans) and testing set (11 to 21 with 20351 scans). SemanticKITTI provides as many as 28 classes, but the official evaluation ignores classes with only a few points and merging classes with different mobility states, thus using a set of 19 valid classes.

\textit{nuScenes}~\cite{caesar2020nuscenes} for LiDAR semantic segmentation is a newly released dataset, with 1000 scenes collected from different areas of Boston and Singapore. Each scene is 20s long and sampled at 20Hz with a Velodyne HDL-32E sensor, thus the nuScenes has 40,000 frames in total. It splits 8130 samples for training, 6019 for validation and 6008 for testing. After merging similar classes and removing rare classes, a total of 16 classes for the LiDAR semantic segmentation are remained.

{\bf Evaluation Metric.}
As official guidance\cite{semantickitti, caesar2020nuscenes} suggests, we use mean intersection-over-union (mIoU) over all classes as the evaluation metric. The mIoU can be formulated as: 
\begin{equation}
mIoU=\frac{1}{C}\sum_{c=1}^C\frac{TP_c}{TP_c+FP_c+FN_c} \label{mIoU}
\end{equation}
where $TP_c$, $FP_c$, $FN_c$ denote true positive, false positive, and false negative predictions for class $c$ and $C$ is the number of classes.

{\bf Network Setup.}
As shown in Figure~\ref{fig:framework}, both range and voxel branch are Unet-like architecture with a stem, four down-sampling and four up-sampling stages, and the dimensions of these 9 stages are $32, 64, 128, 256, 256, 128, 128, 64, 32$, respectively. For range branch, the input range-image size is $64\times 2048$ on SemanticKITTI dataset, and $32\times 2048$ as initial size for nuScenes dataset, then resized to $64\times 2048$ to keep the same as SemanticKITTI. As for voxel branch, the voxel resolution is $0.05m$ for the experiments in Sec.~\ref{sec: results_kitti} and Sec.~\ref{sec: results_nuscenes}. For point branch, it consists of four per-point MLPs with dimensions: $32, 256, 128, 32$. 

{\bf Training and Inference Details.}
We employ the commonly used \textit{cross-entropy} as loss in training.
Our RPVNet is trained from scratch in an end-to-end manner with the ADAM or SGD optimizer. For the SemanticKITTI dataset, the model uploaded to the leaderboard was trained with SGD, batch size 12, learning rate 0.24 for 60 epochs on 2 GPUs, which is kept the same as SPVCNN~\cite{spvnas} for fair comparison. This setup takes around 100 hours on 2 Tesla V100 GPUs. For the other experiments, including nuScenes dataset and ablation studies, we train the entire network with ADAM, batch size 40, learning rate 0.003 for 80 epochs on 8 Tesla V100 GPUs. The cosine annealing learning rate strategy is adopted for the learning rate decay. During training, we utilize the widely used data augmentation strategy of segmentation, including global scaling with a random scaling factor sampled from $[0.95, 1.05]$, and global rotation around the $Z$ axis with a random angle. We also conduct the proposed \textit{instance cut-mixup} sampling strategy to fine-tune the network in the last 10 training epoch.  To note that, in the voxelization process, we set the max number of voxels to $84000$ for training, and all voxels for inference.

\begin{table}[]
\begin{center}
\resizebox{\linewidth}{!}{%
\begin{tabular}{l|ccc|c}
\hline
View            & \#params(M) & \#Macs(G) & latency(ms) & mIoU \\ \hline\hline
RandLA-Net~\cite{hu2020randla}      & 1.2         & 66.5      & 416         & 53.9 \\
SqueezeSegV3~\cite{xu2020squeezesegv3}    & 26.2        & 515.2     & 113         & 55.9 \\
MinkowskiNet~\cite{MinkowskiNet}    & 21.7        & 114       & 139         & 63.1 \\
SPVCNN~\cite{spvnas} & 21.8        & 118.6     & 150         & 63.8 \\ \hline
\textbf{RPVNet}    & 24.8        & 119.5     & 168            &  \textbf{70.3}    \\ 
\textbf{RPV$^{\dag}$Net}    & 24.8        & 74.6     & 111            &  68.3    \\ \hline  
\end{tabular}%
}
\end{center}
\caption{Illustration of the efficiency and performance trade-off of our RPVNet. The mIoU on SemanticKITTI test set is reported. The latency is tested on Tesla V100, including pre- and post-processing time. V$^{\dag}$ indicates the voxel resolution is 0.1m.}
\label{tab:ae-tradeoff}
\end{table}

\subsection{Results on SemanticKITTI}
\label{sec: results_kitti}
In this experiment, we compare the results of our proposed method with existing state-of-the-art LiDAR segmentation methods on SemanticKITTI test set. As shown in Table~\ref{tab:res-kiti}, our method shows superior performance than all existing methods in term of mIoU. Thanks to the strong robustness of our multi-view interactive learning, RPVNet performs best in 11 of 19 categories. From Table~\ref{tab:res-kiti}, we can discover that: 1). The voxel-based method is relatively better than the point- and range-based method, while state-of-the-art method based on the point and range is basically equivalent; 2). The multi-view fusion method is relatively better than voxel-based method.

Despite using the multi-view features, our RPVNet still shows a very competitive running latency due to the compact network design and efficient implementation, as illustrated in Table~\ref{tab:ae-tradeoff}

\subsection{Results on nuScenes}
\label{sec: results_nuscenes}
In this experiment, we report the results of our proposed method on the newly released nuScenes validation set. As shown in Table~\ref{tab:res-nuscenes}, our RPVNet achieves better performance when compared to previous competitive methods, including RangeNet++~\cite{iros19rangenetplusplus}, PolarNet~\cite{zhang2020polarnet}, Salsanext~\cite{cortinhal2020salsanext}, AMVNet~\cite{liong2020amvnet} and Cylinder3D~\cite{zhu2020cylinder3d}, and Salsanext perform the post-processing. Specifically, the proposed method obtains about $5\% \sim 12\%$ performance gain than range-based methods.

\subsection{Ablation Studies}
\label{sec: Ablation Studies}

{\bf Effects of different views.}
This is a core experiment to illustrate the insight of our motivation. We conduct extensive control experiments, including single view(R, P, V), fusion of two views(RP, PV), and fusion of all three views(RPV). To note that, only $1/4$ training data was used in order to speed up the pace of training, and the \textit{mIoU} was reported on SemanticKITTI validation set(sequence 08) and nuScenes validation set, moreover \textit{Macs} and \textit{latency} were tested on SemnaticKITTI dataset. 

From the results in Table~\ref{tab:view_effect}, we can discover that: 1). The RP fusion is effective when compare single R or P with RP; 2). The PV fusion is effective when compare single P or V with PV, especially when voxel resolution is relatively low; 3). The RPV fusion is effective when compare RP or PV with RPV; 4). The proposed RPV fusion can boost performance consistently with an acceptable complexity overhead; 5). RPV fusion can achieve comparable performance even if voxel resolution is low, making it valuable for real-time application.

\begin{table}[]
\begin{center}
\resizebox{\linewidth}{!}{%
\begin{tabular}{l|cc|cc}
\hline
\multirow{2}{*}{Views} &
  \multirow{2}{*}{\begin{tabular}[c]{@{}c@{}}\#params\\ (M)\end{tabular}} &
  \multirow{2}{*}{\begin{tabular}[c]{@{}c@{}}Macs\\ (G)\end{tabular}} &
  \multicolumn{2}{c}{mIoU} \\ \cline{4-5} 
       &       &       & SemanticKITTI & nuScenes      \\ \hline\hline
PV~\cite{spvnas} & 21.8  & 118.6 & 61.5          & 70.6          \\ \hline
R      & 2.62  & 27.5  & 47.5          & 56.7          \\
P      & 0.057 & 5.3   & 14.4          & 16.8          \\
V      & 22.1  & 88.1  & 63.6          & 69.5          \\
V$^{\dag}$      & 22.1  & 43.2  & 60.8          & 72.8          \\
V$^{\ddag}$     & 22.1  & 11.4  & 52.8          & 66.5          \\ \hline
RP     & 2.67  & 31.8  & 50.5          & 59.0          \\
PV     & 22.2  & 92.5  & 63.8          & 71.7          \\
PV$^{\dag}$     & 22.2  & 47.5  & 61.1          & 74.3          \\
PV$^{\ddag}$    & 22.2  & 15.8  & 54.4          & 68.7          \\ \hline
RPV    & 24.8  & 119.5 & \textbf{65.5} & 73.7          \\
RPV$^{\dag}$    & 24.8  & 74.6  & 63.5          & \textbf{74.6} \\
RPV$^{\ddag}$   & 24.8  & 42.8  & 58.4          & 70.2          \\ \hline
\end{tabular}%
}
\end{center}
\caption{Effects of different views. Only 1/4 training data was used for quick experiment. The first row PV is SPVCNN~\cite{spvnas}, and re-trained from scratch with 1/4 data. V, V$^{\dag}$ and V$^{\ddag}$ represent different voxel resolution: V is 0.05m(same as SPVCNN), V$^{\dag}$ is 0.1m, and V$^{\ddag}$ is 0.3m. Macs was computed on SemanticKITTI dataset. The results on both SemanticKITTI and nuScenes consistently show that the fusion pipeline, R$\to$RP$\to$RPV or V$\to$PV$\to$RPV, is effective with acceptable computation overhead.}
\label{tab:view_effect}
\end{table}

\begin{table}[]
\begin{center}
\resizebox{0.75\linewidth}{!}{%
\begin{tabular}{l|cc}
\hline
\multirow{2}{*}{Method} & \multicolumn{2}{c}{mIoU} \\ \cline{2-3} 
                        & SemanticKITTI  & nuScenes \\ \hline\hline
Addition                & 67.2              & 76.9     \\
Concatnation            & 67.4               & 77.1      \\
Gated fusion            & 68.2               & 77.6     \\ \hline
\end{tabular}%
}
\end{center}
\caption{Variants of fusion type. The mIoU on SemanticKITTI validation set without the \textit{instance CutMix} is reported.}
\label{tab:fusion_type}
\end{table}


{\bf Variants of fusion style.}
We investigate the effect of varying the fusion style for multi-view features as shown in Fig.\ref{tab:fusion_type}. We can see that Gated fusion methods obtained improved results by 1\% mIoU in SemanticKITTI and  0.7\% mIoU in nuScenes than Addition.


\begin{table}[]
\begin{center}
\resizebox{0.55\linewidth}{!}{%
\begin{tabular}{l|ccc}
\hline
View & ensemble & fusion & gap \\ \hline\hline
RP   & 49.7     & 54.3   & 4.6 \\
PV   & 63.9     & 64.8   & 0.9 \\
RPV  & 64.5     & 68.2   & 3.7 \\\hline
\end{tabular}%
}
\end{center}
\caption{Comparison with model ensemble. The mIoU on SemanticKITTI validation set is reported.}
\label{tab:ensemble}
\end{table}


{\bf Better than Ensemble?}
It inevitably comes up the question: Is multi-view fusion methods better than model ensemble? We conduct several ablation experiments to illustrate this problem, and the ensemble is conducted by accumulating multi-models' final softmax scores. As shown in Table~\ref{tab:ensemble}, you can see that our fusion method is far better than model ensemble.

\begin{table}[]
\begin{center}
\resizebox{0.9\linewidth}{!}{%
\begin{tabular}{cccc|cc}
\hline
\multirow{2}{*}{\begin{tabular}[c]{@{}c@{}}Baseline\\ V-branch\end{tabular}} &
  \multirow{2}{*}{\begin{tabular}[c]{@{}c@{}}Add\\ P-branch\end{tabular}} &
  \multirow{2}{*}{\begin{tabular}[c]{@{}c@{}}Add\\ R-branch\end{tabular}} &
  \multirow{2}{*}{\begin{tabular}[c]{@{}c@{}}instance\\  cut-mix\end{tabular}} &
  \multicolumn{2}{c}{mIoU} \\ \cline{5-6} 
                     &                      &                      &                      & SemanticKITTI & nuScenes \\ \hline\hline
$\surd$ &       &          &         &   64.2       & 72.3          \\
$\surd$ & $\surd$ &        &       & 64.8     & 74.8         \\
$\surd$ & $\surd$ & $\surd$ &          &  68.2     & \textbf{77.6}          \\
$\surd$ & $\surd$ & $\surd$ & $\surd$ &  \textbf{69.6}     & -      \\ \hline
\end{tabular}%
}
\end{center}
\caption{Effects of overall components. Tested on SemanticKITTI and nuScenes validation set.}
\label{tab:component}
\end{table}


{\bf Effects of overall components.}
Tacking all effective components together, it comes to our final solution. We compare each component's effect in Table~\ref{tab:component}, and as you can see, adding R-branch is more effective than P-branch, and instance cut-mix is also essential for the final performance.


\section{Conclusion}
In this paper, we have proposed a deep and efficient range-point-voxel fusion network, namely RPVNet, for LiDAR point cloud segmentation, where three different views enhance each other in an adaptive way. Specifically, we summarize the efficient multi-view interactive paradigm, including multi-view representation indexing and feature propagation, which gives a solid foundation for interacting point-level context information among multiple views. Finally, We conduct extensive experiments to illustrate the effectiveness and efficiency of RPVNet, and achieves state-of-the-art results on two large-scale public datasets.


\section{Acknowledgements}
This work was supported by National Key R\&D Program of China (Grant No.2020AAA010400X).


{\small
\bibliographystyle{ieee_fullname}
\bibliography{egbib}
}

\end{document}